\documentclass[10pt,twocolumn,letterpaper]{article}

\newif\ifcameraready
\camerareadytrue   %
\ifcameraready
  \usepackage[pagenumbers]{cvpr} %
\else
  \usepackage[review]{cvpr}      %
\fi
\usepackage{cuted}     %
\usepackage{caption}   %

\usepackage{graphicx}
\usepackage{booktabs}
\usepackage{wrapfig}
\usepackage{cuted}
\usepackage{caption}

\definecolor{cvprblue}{rgb}{0.21,0.49,0.74}
\usepackage[pagebackref,breaklinks,colorlinks,allcolors=cvprblue]{hyperref}

\def\confName{CVPR}
\def\confYear{2027}

\title{InfantNet: A Large Scale Dataset for Infant Body Pose and Shape Estimation}

\author{
    Zhipeng Hui\textsuperscript{1,*} \quad
    Jiahao Wang\textsuperscript{1,*} \quad
    Xun Dong\textsuperscript{1,*} \quad
    Guofeng Zhang\textsuperscript{1} \\
    Xingrui Wang\textsuperscript{1} \quad
    Jiawei Peng\textsuperscript{1} \quad
    Qihao Liu\textsuperscript{1} \quad
    Xiaoding Yuan\textsuperscript{1} \quad
    Yi Zhang\textsuperscript{2} \\
    Junjie Oscar Yin\textsuperscript{3} \quad
    Wufei Ma\textsuperscript{1} \quad
    Qinfan Lyu\textsuperscript{1} \quad
    Sara N. Lupolt\textsuperscript{1} \\
    Keeve Edward Nachman\textsuperscript{1} \quad
    Adam Kortylewski\textsuperscript{4} \quad
    Alan Yuille\textsuperscript{1} \\[0.4em]
    {\normalsize
    \textsuperscript{1}Johns Hopkins University \quad
    \textsuperscript{2}Sony Corporation of America \quad
    \textsuperscript{3}University of Washington} \\
    {\normalsize \textsuperscript{4}CISPA Helmholtz Center for Information Security \quad \textsuperscript{*}Equal contribution}
}

\begin{document}
\maketitle
\ifcameraready\def\teaserwidth{0.67}\else\def\teaserwidth{0.88}\fi
\begin{strip}
  \centering
  \includegraphics[width=\teaserwidth\linewidth]{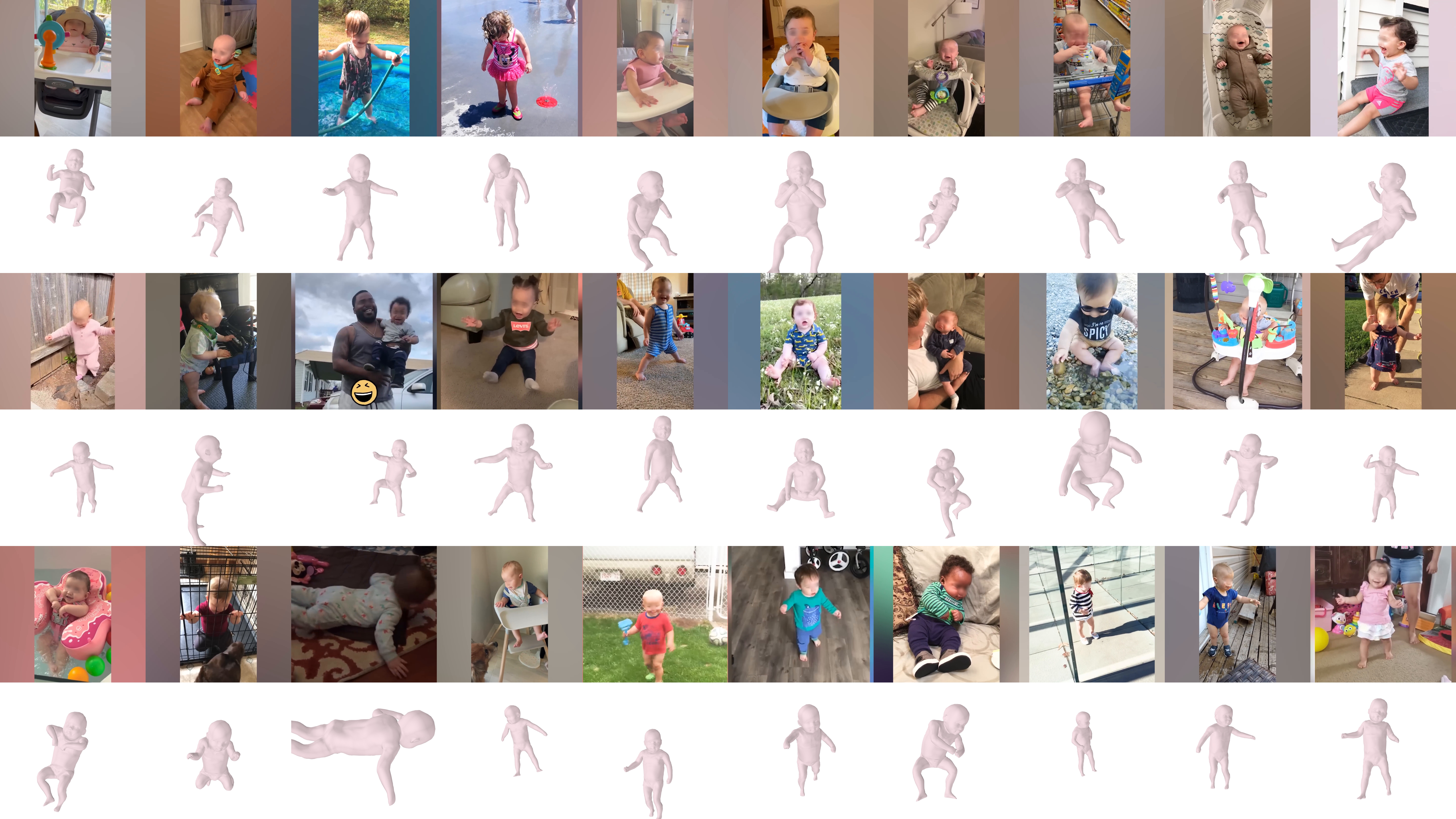}
  \captionof{figure}{Examples of SMIL meshes from InfantNet. The dataset captures diverse infant morphology, articulation, and real-world conditions, providing reliable large-scale 3D supervision for training and testing infant-specific models.}
  \label{fig:smil_sample_results}
\end{strip}

\begin{abstract}
Infant pose and shape estimation is essential for applications in childcare, developmental monitoring, and medical diagnosis. However, existing methods and datasets are largely designed for adults, and direct transfer to infants fails due to substantial differences in body proportions, articulation limits, and frequent self-occlusion. To address this gap, we introduce InfantNet, the largest real-image infant dataset to date, comprising 108,902 RGB images of infants aged 6-18 months. Each image is annotated with 2D keypoints, and a curated subset of 11,642 images additionally includes 3D pose and shape annotations with full SMIL parameters, verified through multi-view rendering and reprojection-error filtering. We use an iterative annotation pipeline to ensure high fidelity across both 2D and 3D labels. InfantNet establishes a large-scale, comprehensive benchmark for infant 2D keypoint detection and 3D pose-and-shape recovery. Baseline experiments demonstrate that state-of-the-art adult pose estimators do not generalize well to infants, whereas fine-tuning on InfantNet yields a consistent improvement. The gains are even more pronounced for 3D pose and shape estimation, where SMIL-based estimation demonstrates robust performance with accuracy comparable to results reported for SMPL-based models. By releasing the InfantNet dataset and benchmark, we provide a vital resource for advancing infant pose analysis and related healthcare applications.
\end{abstract}
    
\section{Introduction}
\label{sec:intro}

Estimating the pose and shape of the human body is a foundational task in computer vision, with broad applications in action recognition, human-computer interaction, and healthcare~\cite{kanazawa2018end, kolotouros2019learning, pavlakos2019expressive, lupolt2025development, lupolt2025application}. Over the past decades, human pose and shape estimation algorithms have achieved remarkable progress. However, existing methods and datasets predominantly focus on adults, whose body morphology and poses differ significantly from those of infants. Directly applying adult-trained models to infants results in poor performance due to substantial domain gaps, including differences in body proportions, limited articulation, and frequent self-occlusion typical of infant behavior. %
These limitations significantly hinder the deployment of such models in infant-related applications, despite their critical importance in areas such as developmental monitoring, childcare, and early medical diagnosis.
For instance, quantifying infant micro-activities~\cite{lupolt2025application} requires fine-grained 3D pose analysis. However, in the absence of accurate infant-specific models, researchers~\cite{lupolt2025development} currently rely on keypoint detectors trained primarily on adult data combined with multi-view triangulation as a noisy proxy, which are complicated to deploy in real-world home environments.

A major obstacle to advancing infant-specific pose estimation is the lack of large-scale, high-quality annotated datasets. Existing infant datasets typically contain only a few thousand images~\cite{hesse2018computer,huang2021infant,dechemi2021babynet}, in contrast to adult datasets such as 3DPW~\cite{vonMarcard2018recovering} and Human3.6M~\cite{ionescu2014human36m}, which comprise over 51,000 and 3.6 million images, respectively. Moreover, current infant datasets are often synthetic or focus mainly on sparse 2D annotations~\cite{hesse2018computer, huang2021infant, dechemi2021babynet}, lacking the 3D supervision needed to train and evaluate modern human modeling methods~\cite{xu2022vitpose,goel2023humans}.

To address this gap, we introduce \textbf{InfantNet}, a large-scale dataset specifically designed for infant pose and shape estimation. Comprising more than 100K high-resolution RGB images of infants mainly aged 6 to 18 months, InfantNet includes accurate annotations of 17 anatomically consistent 2D keypoints. Frames are sampled from videos at up to 2 FPS to reduce temporal redundancy while preserving pose diversity (Fig.~\ref{fig:part_diversity}). A carefully curated subset of 11,642 images further includes 3D pose and shape annotations by SMIL~\cite{SMIL} parameters, enabling future research for 3D infant shape and pose estimation. We adopt a robust iterative annotation pipeline that combines human supervision with optimization-based refinement to ensure high-quality labels in 2D. We further validate 3D SMIL~\cite{SMIL} parameters by predicting the resulting meshes from multiple viewpoints and computing reprojection errors. Examples of images and 3D pose and shape annotations are shown in Figure~\ref{fig:smil_sample_results}. 

InfantNet is designed with a \textbf{``privacy-first'' data collection framework}. The HomeVisit subset was collected under institutional review board (IRB) approval with written informed caregiver consent and digital face-masking. For the YouTube subset, we strictly adhere to a non-redistribution policy, releasing only original URLs and anonymized derivatives (keypoints, SMIL parameters). To prevent misuse, the dataset is restricted to research via controlled academic licensing, a signed Data Use Agreement (DUA) prohibiting biometric identification, and a robust takedown mechanism. (See Sec.~\ref{subsec:ethics} for details.)

Our benchmark shows that state-of-the-art adult pose estimators generalize poorly to infants. Fine-tuning on InfantNet substantially reduces this gap, improving 2D performance by 6.2\% AP and reducing 3D MPJPE by 78\%. Despite training on only 10,071 frames, our infant-specific 4D-Humans model achieves 46.7 mm PA-MPJPE on InfantNet, approaching the adult result of 32.4 mm~\citep{goel2023humans} trained and tested on the Human3.6M dataset, and surpasses HMR (56.8 mm) and PARE (50.6 mm). These results highlight the strong data efficiency of InfantNet, enabling high-fidelity 3D modeling with $350\times$ fewer training samples than typical adult datasets. Our model also enables significantly more accurate detection of infant hand-to-mouth behavior~\cite{lupolt2025development}, supporting reliable assessment of environmental exposure risks in real-world settings.

In summary, the main contributions of our paper are:

\begin{itemize}
    \item We introduce \textbf{InfantNet}, a large-scale dataset for infant pose and shape estimation, containing 108,902 RGB images with high-quality 2D keypoint annotations. A curated subset of 11,642 images further includes 3D pose and shape labels based on SMIL. The dataset enables comprehensive evaluation of both 2D and 3D pose and shape estimation models.

    \item We develop an end-to-end SMIL-based framework for infant 3D pose and shape estimation. To our knowledge, this is the first attempt to directly regress SMIL parameters from images with neural networks. By generating high-quality pseudo ground-truth through optimization and verification, we train end-to-end models to directly regress SMIL parameters from images, enabling accurate infant-specific 3D modeling.

    \item We systematically analyze the domain gap between adult and infant pose estimation. Fine-tuning on InfantNet substantially improves both 2D and 3D performance, demonstrating the importance of domain-specific data. Our models achieve competitive accuracy with far fewer training samples and support downstream applications such as infant microactivity analysis.

    \item We set up a benchmark on InfantNet using state-of-the-art human pose and shape estimation algorithms. 

\end{itemize}

By releasing data, code, and evaluation protocols, we hope to provide a foundation for future research in infant pose and shape estimation, with implications for early developmental assessment, clinical monitoring, and beyond. 

\section{Related Work}
\label{sec:related-work}
In this section, we discuss existing infant datasets and related pose and shape estimation algorithms.

\begin{table}[t]
\centering
\footnotesize
\setlength{\tabcolsep}{4pt}
\resizebox{\linewidth}{!}{%
\begin{tabular}{lccccc}
\toprule
\textbf{Dataset} & \textbf{Syn/Real} & \textbf{\# Images} & \textbf{Keypoints} & \textbf{Segmentation} & \textbf{3D Annotations} \\
\hline
BabyPose~\cite{migliorelli2020babypose} & Real & 16,000 & Yes & No & Depth only \\

BabyNet~\cite{dechemi2021babynet} & Real & 2,984 & No & BBox only & No \\

MINI-RGBD~\cite{hesse2018computer} & Synthetic & 12,000 & Yes & No & Depth only \\

SyRIP~\cite{huang2021infant} & Syn+Real & 1,700 & Yes & No & No \\

\textbf{InfantNet (ours)} & Real & 108,902 & Yes & Yes & Yes \\
\bottomrule
\end{tabular}
}
\caption{
Comparison of InfantNet with existing open-source infant pose datasets. InfantNet is the largest real-world infant dataset to date, providing 108,902 images with keypoint annotations, full segmentation masks, and a curated subset with 3D SMIL parameters. In contrast, prior datasets are limited in scale, modality, or 3D supervision (e.g., MINI-RGBD provides depth only).
}
\label{tab:related}
\end{table}

\subsection{Infant Datasets}
Human pose and shape estimation has benefited enormously from large-scale adult datasets such as Human3.6M~\cite{ionescu2014human36m}, MPII~\cite{andriluka20142d}, and AMASS~\cite{mahmood2019amass}. In contrast, infant datasets remain scarce due to ethical, logistical, and technical challenges in capturing infant motion in naturalistic environments. Early efforts include BabyPose~\cite{migliorelli2020babypose}, which provides 16 depth videos from NICU settings with 12-joint annotations, and the dataset from~\cite{groos2022towards}, focusing on supine infants with limited poses. BabyNet~\cite{dechemi2021babynet} targets activity classification without fine-grained pose labels. MINI-RGBD~\cite{hesse2018computer} offers synthetic depth sequences of 12 infant models, while SyRIP~\cite{huang2021infant} combines 1,700 synthetic and real images with sparse 2D keypoints. These datasets are small, often synthetic or modality-specific, and lack segmentation masks or 3D ground truth pose and shape parameters.
Our proposed InfantNet addresses these limitations by providing real infant RGB images with high-quality 2D keypoints, segmentation masks, bounding boxes, and 3D pose and shape annotations. Table~\ref{tab:related} compares InfantNet with prior datasets, showing its unique scale, diversity, and multimodal supervision, which together enable robust evaluation and training of infant-specific models.

\subsection{Infant Pose and Shape Estimation}
2D infant pose estimation initially applied general-purpose models such as OpenPose~\cite{cao2019openpose}, HRNet~\cite{sun2019deep}, and AlphaPose~\cite{fang2022alphapose}, but these perform poorly due to infants’ distinct proportions, articulation limits, and frequent self-occlusion~\cite{sciortino2017estimation}. Later work fine-tuned these CNN-based estimators on small infant datasets~\cite{groos2022towards}, reducing error and approaching human inter-rater variability. SHIFT~\cite{bose2025shift} adapts adult data to unlabeled infant data, reflecting the scarcity of labeled infant-specific datasets. More recent transformer-based methods, such as ViTPose~\cite{xu2022vitpose}, AggPose~\cite{cao2022aggpose}, and Sapiens~\cite{khirodkar2024sapiens}, improve spatial reasoning in cluttered settings~\cite{jahn2025comparison}, but are susceptible to overfitting without large and diverse infant-specific datasets.

In the 3D domain, early infant methods relied on depth sensors for direct joint prediction~\cite{hesse2018computer} or stereo systems for triangulation~\cite{soualmi20243d}, achieving good accuracy in clinical settings but requiring specialized hardware. The SMIL model~\cite{SMIL} represents a key step toward parametric modeling of infants: adapting the adult SMPL template~\cite{loper2015smpl} with an infant-specific mesh and priors learned from noisy, incomplete RGB-D recordings. While SMIL enables full-body mesh registration, its reliance on limited and low-quality depth sequences constrains generalization. ZeDO-i~\cite{zhou2024zedo} adapts adult priors with limited infant data, focusing on keypoint estimation. Other approaches leverage synthetic renderings~\cite{huang2021infant} or alternative sensors such as pressure mats~\cite{donati2013modular,clever2020bodies}, but these are less practical in daily settings.

Despite this progress, the field lacks a large-scale, real-world RGB benchmark with consistent 2D and 3D annotations. Our \textit{InfantNet} fills this gap, enabling systematic development and evaluation of infant pose and shape estimation models that generalize across naturalistic scenarios.

\subsection{Adult pose and shape estimation}
3D human pose and shape estimation has been extensively studied through parametric models such as SMPL~\cite{loper2015smpl} and its extensions (e.g. SMPL-X~\cite{pavlakos2019expressive}). Learning-based approaches including HMR~\cite{kanazawa2018end}, VIBE~\cite{kocabas2020vibe}, PARE~\cite{Kocabas2021PARE}, and 4D-Humans~\cite{goel2023humans} use feedforward regression to directly predict body parameters from images or videos, often with strong temporal consistency. However, these methods are trained on adult datasets and exhibit poor generalization to infants, whose body proportions, articulation limits, and frequent occlusions differ significantly. In this work, we adapt such adult regression networks to the infant domain using InfantNet and SMIL-based annotations.

\section{InfantNet Dataset}
\label{sec:method}

In this section, we present the InfantNet dataset and detail the data collection and annotation procedures.

\subsection{Ethical Design and Data Governance}
\label{subsec:ethics}
The collection, processing, annotation, and release of InfantNet follow 
the IEEE Code of Ethics, with a multi-layer ethical protocol to ensure privacy.

\noindent\textbf{IRB and Consent.} The HomeVisit subset was collected under an approved Institutional Review Board (IRB) protocol. We obtained written informed consent from all caregivers, explicitly covering anonymized data sharing for research.

\noindent\textbf{Privacy and Anonymization.} To protect identities, we apply digital face-masking to all YouTube and HomeVisit frames and strip all identifying metadata. For the YouTube subset, we strictly follow a non-redistribution policy: we release only anonymized, derived data (keypoints, SMIL parameters) and original URLs. Figure~\ref{fig:smil_sample_results} shows our face-masking strategy, and more examples can be found in Figure~\ref{fig:3d_annotations_additional_samples} of Appendix~\ref{sec: 3d_annotations}. We also provide experimental results that demonstrate the robustness of our model and viability of our dataset under this strategy in Appendix~\ref{sec: 3d_face_masking}.

\noindent\textbf{Controlled Access.} 
InfantNet will be released on Hugging Face with gated and manually approved access under CC BY-NC-ND 4.0. Access requires institutional verification and agreement to a Data Use Agreement prohibiting commercial use, biometric identification, and redistribution of modified derivatives. A dedicated takedown mechanism is maintained for ongoing compliance.

\noindent\textbf{Societal Impact.} InfantNet addresses the lack of large-scale resources for infant healthcare research. Our goal is to enable better care for infants while maintaining the highest standards of participant privacy. Specific notes on data and code publications can be found in Appendix~\ref{sec: public_release}.

\subsection{Data Collection}
\noindent{\textbf{YouTube data collection. } }
We curated publicly available infant videos from YouTube, capturing infants in a wide range of poses and body shapes. These raw videos serve as one source of our dataset and are later processed through annotation and quality filtering. The YouTube data is collected strictly following the platform policy and privacy safeguards, as detailed in Sec.~\ref{subsec:ethics}.

\noindent\textbf{HomeVisit Data Collection.} Following written informed consent from all caregivers, we recorded multi-view videos of infants in their domestic play areas. The setup used four statically mounted GoPro Hero9 cameras positioned orthogonally for overlap coverage. To support accurate SMIL fitting, we performed rigorous per-session camera calibration. For HomeVisit, the female-to-male video ratio is 47\%:53\%. We do not infer race, ethnicity, geographic origin, or precise age when such metadata is unavailable.

\subsection{Data Annotation}
\begin{figure*}[t]
  \centering
  \includegraphics[width=\linewidth]{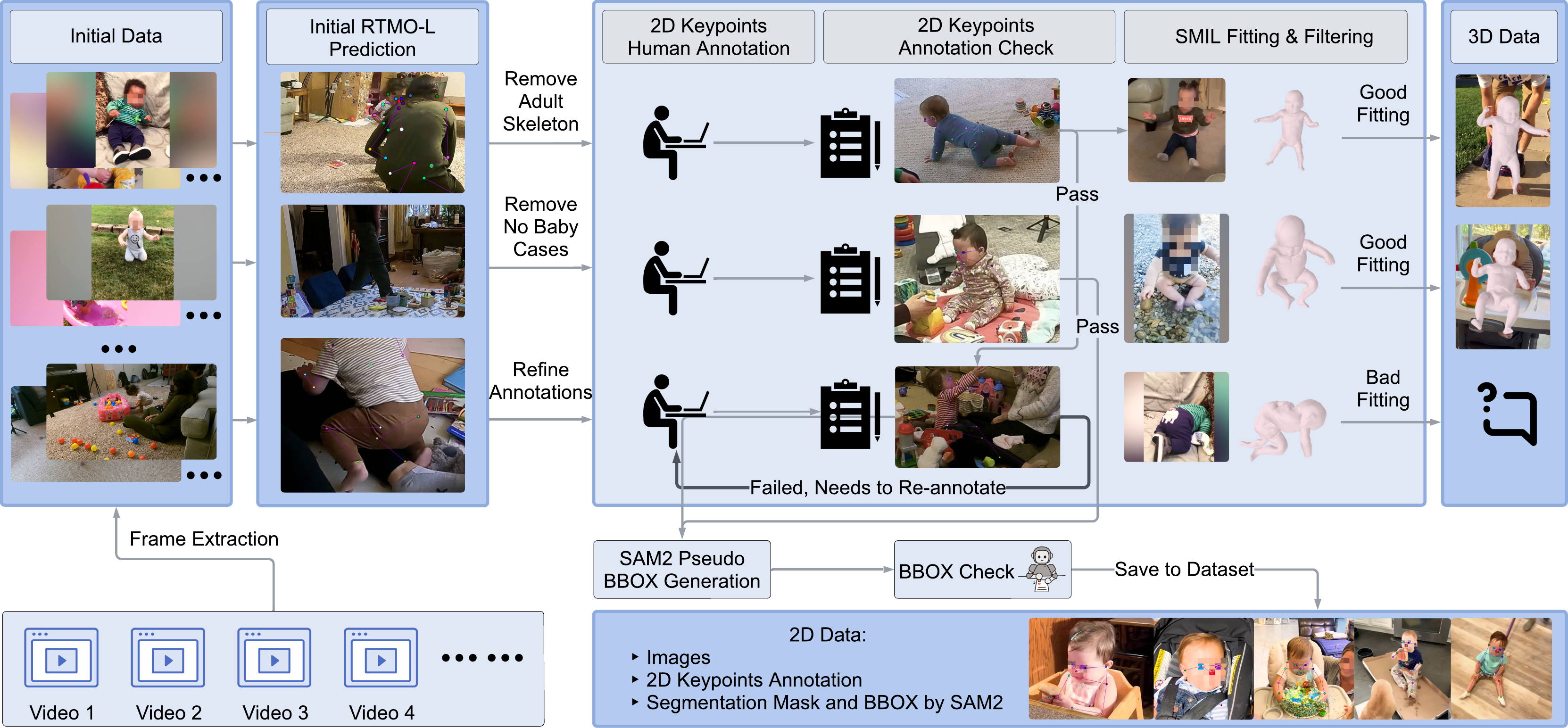}
  \caption{Annotation pipeline of our InfantNet Dataset. Frames from YouTube and HomeVisit recordings were annotated using a semi-automatic, inspector-guided workflow. For 2D labels, RTMO-L provided initial keypoints that were refined by annotators with iterative quality control, following the COCO format. For 3D labels, we applied an optimization-based method to fit SMIL to infant images, and annotators further filtered the results to ensure accurate pose and shape annotations.}
  \label{fig:smil_pipeline}
  \vspace{-5px}
\end{figure*}

We annotated frames from both sources, filtering to ensure exactly one infant per frame and sampling at a maximum rate of 2 FPS to preserve temporal diversity.
In total, 206{,}436 frames were extracted prior to quality filtering. 
The YouTube subset contains 51,039 frames. These videos feature around 25 distinct infants per 4 minutes, yielding an estimated total of around 3,000 unique infants.
The HomeVisit subset consists of 15 multi-camera recording sessions (each with four synchronized cameras), from which 155,397 frames were extracted. 
After frame extraction, we performed 2D body keypoint annotation followed by multi-stage quality control, removing frames with severe occlusion, motion blur, or other degradations. The resulting high-quality annotations serve as ground-truth supervision for SMIL fitting and subsequent model training.

To annotate such a large and heterogeneous dataset, we adopted a semi-automatic, inspector-guided annotation pipeline, inspired by human-in-the-loop strategies in 3D model fitting works such as Animal3D~\cite{xu2023animal3d}. As shown in Figure~\ref{fig:smil_pipeline}, our annotation workflow consisted of six stages to ensure efficiency and accuracy.

\textbf{Initial keypoint estimation.}
We employed RTMO-L~\cite{lu2024rtmo} from MMPose~\cite{mmpose2020} to produce 2D pseudo-labels for all frames. These predictions were loaded into CVAT~\cite{cvat}, where human annotators conducted multi-step refinement.

\textbf{Initial filtering and manual correction.}
Annotators reviewed the RTMO-L~\cite{lu2024rtmo} predictions in CVAT and removed frames with severe motion blur, camera shake, extreme lighting artifacts (e.g., overexposure or harsh shadows), strong occlusion (e.g., when the infant is mostly outside the field of view or heavily covered by objects), missing infants, or subjects older than 18 months. Adult skeletons were discarded, and one infant per frame was retained. For remaining frames, if the RTMO-L prediction provided accurate and anatomically plausible keypoints, the annotation was accepted without modification to maximize efficiency. Otherwise, annotators manually corrected the predictions to ensure anatomical consistency. All annotations followed the COCO format with 17 joints: nose, left eye, right eye, left ear, right ear, left shoulder, right shoulder, left elbow, right elbow, left wrist, right wrist, left hip, right hip, left knee, right knee, left ankle, and right ankle. Each keypoint was tagged as \textit{visible}, \textit{partially occluded}, or \textit{fully occluded}.

\textbf{Inspector verification.} An inspector reviewed annotations for consistency and quality. Frames with annotation errors were returned to annotators with revision feedback.

\textbf{Iterative refinement.} This correction–verification cycle was repeated until the inspector approved all annotations per recording session, ensuring high-quality labels.

\textbf{Bounding box generation.} Once the keypoints were finalized, bounding boxes were generated using SAM2~\cite{ravi2024sam}. Visible keypoints were used to prompt mask generation, and the final bounding boxes were computed as the minimal rectangles covering both the masks and any occluded joints, thereby mitigating segmentation errors due to occlusion (e.g., infants partially covered by furniture).

\begin{figure}[t]
  \centering
  \includegraphics[width=\linewidth]{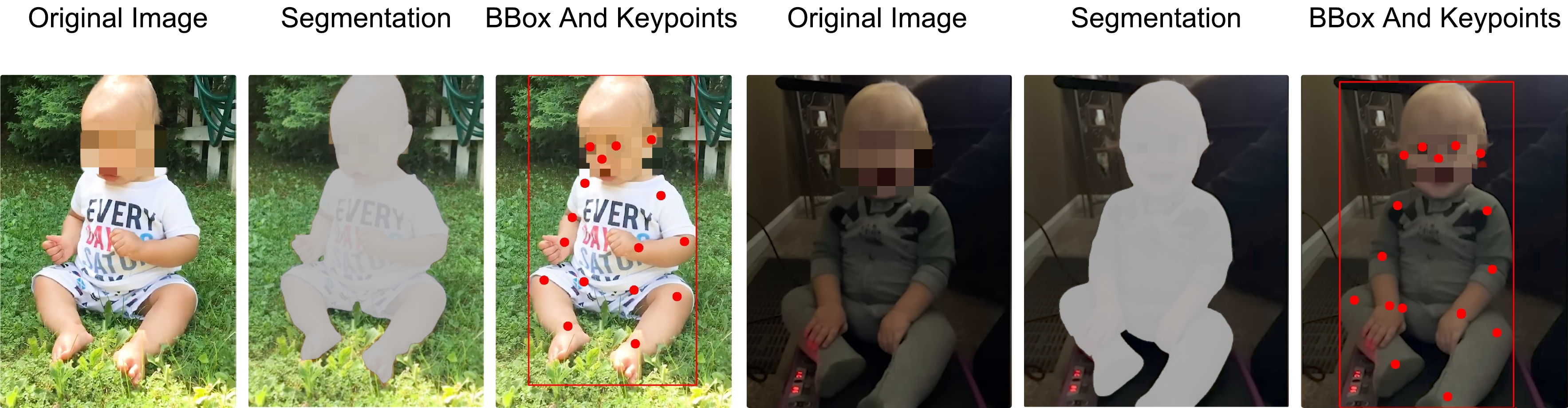}
  \caption{Visualization of different annotation stages in our InfantNet Dataset: Original image, Segmentation mask, Bounding box, and Annotated Body Keypoints.}
  \label{fig:annotation_stages}
  \vspace{-10px}
\end{figure}

\textbf{Final bounding box validation.} Inspectors performed a second review to verify bounding box accuracy and tightness, ensuring annotation completeness and quality.

This staged workflow allowed us to balance scalability and precision, producing high-quality 2D annotations with explicit human oversight at each critical step. Examples of the 2D annotations in different stages are shown in Figure~\ref{fig:annotation_stages}.

\subsection{Fitting SMIL on Infant Images}
\label{subsec:3d_fitting}
We use an optimization-based method~\cite{bogo2016keep,yin2025easyret3d} to fit SMIL~\cite{SMIL} on infant images from our dataset following the FiDIP~\cite{huang2021infant} workflow. We optimize the shape parameters $\beta$ and pose parameters $\theta$ in SMIL by minimizing the loss:
\begin{equation}
    \mathcal{L}_{\text{SMIL}} = \mathcal{L}_{\text{data}} + \mathcal{L}_{\text{pose}} + \mathcal{L}_{\text{shape}} + \mathcal{L}_{\text{others}}.
\end{equation}

The data term, $\mathcal{L}_{\text{data}}$, includes mesh-to-scan $\mathcal{L}_{\text{m2s}}$ and scan-to-mesh $\mathcal{L}_{\text{s2m}}$ distances. $\mathcal{L}_{\text{pose}}$ and $\mathcal{L}_{\text{shape}}$ enforce priors on $\theta$ and $\beta$, respectively. $\mathcal{L}_{\text{others}}$ contains losses for landmark keypoints alignment ($\mathcal{L}_{\text{lm}}$), table contact ($\mathcal{L}_{\text{table}}$), temporal smoothness ($\mathcal{L}_{\text{sm}}$), and self-intersections ($\mathcal{L}_{\text{sc}}$).

The registration proceeds in three steps: estimating initial shape, solving per-frame poses with smoothness and data terms, and registering meshes via constrained deformation. We validate SMIL parameters with multi-view rendering and reprojection checks, removing results inconsistent with the image, distorted in any view, or showing implausible poses or shapes. We also project 3D keypoints to 2D and discard frames with reprojection error above 16 pixels. The filtered parameters provide high-quality 3D ground-truth for end-to-end training. Further 3D annotation quality analysis is provided in Sec.~\ref{subsec:data_summary}.

\subsection{Data Summary}
\label{subsec:data_summary}
We partition the dataset into training and testing splits while maintaining identity consistency across videos and sessions. 
Table~\ref{tab:dataset_summary} reports the statistics of raw frame counts and the final numbers after annotation-stage filtering.

\begin{table}[t]
\centering
\setlength{\tabcolsep}{12pt}
\resizebox{\linewidth}{!}{%
\begin{tabular}{lccc}
\toprule
\textbf{Subset} & \textbf{Viewpoint Info} & \textbf{Raw Frames} & \textbf{Filtered Frames} \\
\midrule
YouTube (Train)   &  Varied & 44,257  & 36,185 \\
YouTube (Test)     &  Varied & 6,782   & 3,959  \\
HomeVisit (Train)   &  4 fixed cameras &123,876 & 56,968  \\
HomeVisit (Test)     &  4 fixed cameras & 31,521  & 11,790 \\
\midrule
\textbf{Total}  & -- & 206,436 & 108,902 \\
\bottomrule
\end{tabular}
}
\caption{Dataset Statistics. ``Filtered Frames'' denotes the final samples retained after removing low-quality, ambiguous, or non-compliant frames during annotation.}
\label{tab:dataset_summary}
\vspace{-10px}
\end{table}

\noindent\textbf{YouTube Subset.}
The YouTube subset consists of \textbf{40,144} frames with diverse environments and camera conditions. \textbf{3,959} frames were designated for testing, while \textbf{36,185} frames were assigned to training, ensuring an identity-disjoint split with no overlap of infants between sets.

\noindent\textbf{HomeVisit Subset.} The HomeVisit subset contains 15 in-home recording sessions, producing a total of 155,397 frames. For evaluation, we selected 3 sessions as the testing set, while the remaining 12 sessions were assigned to training. 
After annotation-stage filtering, the testing set contains \textbf{11,790} frames and the training set contains \textbf{56,968} frames. 

\noindent\textbf{Format.} All annotations follow the standard \textbf{COCO}~\cite{lin2014microsoft} keypoints format and include 17 body keypoints with visibility flags (visible, partially occluded, fully occluded), as well as bounding boxes. The dataset's multi-source and multi-view composition enables both robust training and principled benchmarking in this challenging domain.

\begin{wrapfigure}{r}{0.38\linewidth}
  \vspace{-12pt}
  \centering
  \includegraphics[width=\linewidth]{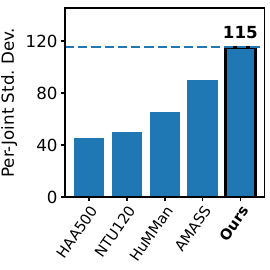}
  \vspace{-19pt}
  \captionof{figure}{
  Pose diversity statistics measured by per-joint standard deviation between InfantNet and large-scale human datasets~\cite{lin2023motionx, haa500, liu2020ntu, cai2022humman, mahmood2019amass}. InfantNet exhibits broad variation in 3D joint configurations.
  }
  \label{fig:part_diversity}
  \vspace{-12pt}
\end{wrapfigure}

\noindent\textbf{Pose Diversity.} We measure pose diversity as the average standard deviation of pelvis-centered 3D joint coordinates across the dataset, capturing spatial variability of articulated body parts, including joint distribution, pose spread, and limb movement range~\cite{lin2023motionx}. The 3D annotations achieve a pose diversity score of \textbf{115.0}, exceeding several widely used large-scale human datasets (Fig.~\ref{fig:part_diversity}) and demonstrating substantial structural variability despite focusing on infants. Prior to multi-view rendering and reprojection-based filtering (Sec.~\ref{subsec:3d_fitting}), the pose diversity score is 138.2. The post-filtering diversity remains at a high magnitude, confirming that the filtering process preserves pose variability and does not bias the dataset toward specific poses or shapes.

\noindent\textbf{Occlusion.} \textbf{75.58\%} of frames have fully occluded keypoints, and \textbf{92.24\%} have fully/partially occluded keypoints. We further categorize occluded keypoints as out-of-frame, self-occluded by the infant body, or externally occluded by other objects. Among occluded keypoints, \textbf{5.5\%} are out of frame, \textbf{64.5\%} are self-occluded, and \textbf{30.0\%} are externally occluded. Additional evaluation and analysis of occlusion and edge cases are provided in Appendix~\ref{sec: edge_cases}.

\noindent\textbf{3D Annotation Quality.} We validate the 3D annotations from consistency and reliability. We assess consistency using multi-view rendering, reprojection error, and bbox-normalized PCK. The 3D annotations achieve a mean reprojection error of \textbf{9.10} px and \textbf{83.04\%/98.40\%/99.99\% PCK@0.05/0.1/0.2}. We evaluate reliability using multi-view joint optimization on calibrated HomeVisit frames. We jointly optimize each body annotation across views using multi-view reprojection error and SMIL priors. The optimized annotations differ from initial annotations by \textbf{80.6 mm} MPJPE, while independently optimized candidates agree closely with only \textbf{3.1 mm} mean pairwise MPJPE, supporting 3D annotation reliability.

\section{Experiments}
\label{sec:experiments}
We benchmark state-of-the-art 2D and 3D pose estimation models on \textbf{InfantNet} to quantify the domain gap between adult-centric models and infant morphology. Secs.~\ref{subsec:2d_keypoint_estimation} and~\ref{subsec:3d_estimation} evaluate 2D transferability and 3D recovery, while Sec.~\ref{subsec:application} shows InfantNet's value in real-world applications.
\subsection{2D Keypoint Estimation}
\label{subsec:2d_keypoint_estimation}
We benchmark a series of widely used 2D human pose estimation models on our proposed infant dataset to evaluate their ability to transfer from adult-centric data to the unique morphology and movement patterns of infants.

\begin{table}[t]
\centering
\setlength{\tabcolsep}{4pt}
\resizebox{\linewidth}{!}{%
\begin{tabular}{l|cccc|cccc}
\toprule
\textbf{Model} & \multicolumn{4}{c|}{\textbf{Direct Inference (\%)}} & \multicolumn{4}{c}{\textbf{Fine-tuned (\%)}} \\
\cmidrule(lr){2-5} \cmidrule(lr){6-9}
 & \textbf{AP} & \textbf{AP.5} & \textbf{AP.75} & \textbf{AR} & \textbf{AP} & \textbf{AP.5} & \textbf{AP.75} & \textbf{AR} \\
\midrule
Hourglass 256x256 & 86.27 & 95.58 & 92.37 & 88.43 & 92.63 & 98.99 & 97.90 & 93.95 \\
Hourglass 384x384 & 87.39 & 96.59 & 93.43 & 89.27 & 93.40 & 98.98 & 97.89 & 94.52 \\
HRFormer-small 256x192 & 87.65 & 96.53 & 93.36 & 89.82 & 93.14 & 98.97 & 97.87 & 94.36 \\
HRFormer-small 384x288 & 88.00 & 96.54 & 94.30 & 90.07 & 93.51 & 98.97 & 97.91 & 94.60 \\
HRFormer-base 256x192 & 88.24 & 95.50 & 93.33 & 90.26 & 93.85 & 98.98 & 97.92 & 95.09 \\
HRFormer-base 384x288 & 88.71 & 96.38 & 94.16 & 90.82 & 94.01 & 98.98 & 97.93 & 95.27 \\
Res50 256x192 & 86.48 & 96.62 & 93.37 & 88.59 & 92.73 & 99.00 & 97.91 & 94.04 \\
Res50 384x288 & 86.57 & 96.58 & 93.31 & 88.63 & 93.03 & 99.00 & 97.93 & 94.16 \\
Res101 256x192 & 87.28 & 96.56 & 93.44 & 89.29 & 93.18 & 98.99 & 97.90 & 94.37 \\
Res101 384x288 & 87.59 & 96.50 & 93.38 & 89.57 & 93.51 & 98.99 & 97.92 & 94.78 \\
Res152 256x192 & 87.31 & 95.56 & 93.41 & 89.38 & 93.31 & 98.98 & 97.92 & 94.57 \\
Res152 384x288 & 87.89 & 96.46 & 93.33 & 89.99 & 93.87 & 98.95 & 97.89 & 94.90 \\
HRNet-w32-dark 256x192 & 88.60 & 96.61 & 93.48 & 90.37 & 93.82 & 98.98 & 97.92 & 94.89 \\
HRNet-w32-dark 384x288 & 88.74 & 96.50 & 94.29 & 90.64 & 93.98 & 98.97 & 97.92 & 95.19 \\
HRNet-w48-dark 256x192 & 88.92 & 96.56 & 94.40 & 90.69 & 94.16 & 98.96 & 97.91 & 95.28 \\
HRNet-w48-dark 384x288 & 88.74 & 96.51 & 94.32 & 90.60 & 94.21 & 98.98 & 97.91 & 95.34 \\
HRNet-w32 256x192 & 88.04 & 96.60 & 93.45 & 89.95 & 93.36 & 98.98 & 97.88 & 94.50 \\
HRNet-w32 384x288 & 88.19 & 96.53 & 93.39 & 90.24 & 93.64 & 98.96 & 97.88 & 94.82 \\
HRNet-w48 256x192 & 88.38 & 96.57 & 93.41 & 90.24 & 93.74 & 98.98 & 97.92 & 94.85 \\
HRNet-w48 384x288 & 88.37 & 96.50 & 93.37 & 90.39 & 93.97 & 98.97 & 97.90 & 95.12 \\
ViTPose-small 256x192 & 87.56 & 96.56 & 93.47 & 89.46 & 93.36 & 98.99 & 97.93 & 94.43 \\
ViTPose-base 256x192 & 89.02 & 96.54 & 94.48 & 90.89 & 93.91 & 98.97 & 97.91 & 95.11 \\
ViTPose-large 256x192 & 90.02 & 96.44 & 94.34 & \textbf{91.99} & 94.75 & 98.96 & 97.91 & 95.81 \\
ViTPose-huge 256x192 & 89.95 & 96.39 & 94.30 & \textbf{91.99} & \textbf{94.77} & 98.95 & 97.90 & \textbf{95.82} \\
Sapiens-0.3B & \textbf{90.08} & 97.60 & 95.43 & 91.96 & 93.35 & 98.97 & 97.90 & 94.48\\
\bottomrule
\end{tabular}
}
\caption{
Comparison of 2D pose estimation models on the InfantNet benchmark under two settings: direct inference using pretrained models and fine-tuning on InfantNet. We report Average Precision (AP), AP at IoU thresholds 0.5 (AP.5) and 0.75 (AP.75), and Average Recall (AR). Direct transfer of pretrained models leads to a clear performance drop, while fine-tuning consistently improves performance, with an average relative gain of \textbf{6.2\% in AP} and \textbf{5.2\% in AR}. ViTPose-huge achieves the best results after adaptation (AP: \textbf{94.77\%}, AR: \textbf{95.82\%}), highlighting the importance of domain-specific data for accurate infant pose estimation.
}
\label{tab:infantnet_results}
\vspace{-10px}
\end{table}

\noindent\textbf{Setup.}
We benchmark Transformer-based models (ViTPose~\cite{xu2022vitpose}, HRFormer~\cite{yuan2021hrformer}, Sapiens~\cite{khirodkar2024sapiens}), CNN-based models (HRNet~\cite{sun2019deep}, Hourglass~\cite{newell2016stacked}), and CNN variants with DARK post-processing (HRNet-Dark~\cite{zhang2020distribution}). All models are initialized from COCO-pretrained weights. We compare models with and without fine-tuning on our InfantNet train split ($\text{YouTube}_{\text{Train}} \cup \text{HomeVisit}_{\text{Train}}$) for 40 epochs following the MMPose setting, with a learning rate of \(1 \times 10^{-4}\) on 4 NVIDIA A100-SXM4-40GB GPUs. 

\noindent\textbf{Evaluation.}
We select the best-performing checkpoint on the validation set and report its corresponding results.
The split preserves identity separation across sessions.
Metrics follow COCO's 17-keypoint protocol: mean Average Precision (AP), AP.5, AP.75, and Average Recall (AR).

\begin{table}[t]
\centering
\setlength{\tabcolsep}{8pt}
\resizebox{\linewidth}{!}{%
\begin{tabular}{l|cc|cc}
\toprule
\textbf{Model} & \multicolumn{2}{c|}{\textbf{Adult Body Model}} & \multicolumn{2}{c}{\textbf{Infant Body Model}} \\
\cmidrule(lr){2-3} \cmidrule(lr){4-5}
 & \textbf{MPJPE$\downarrow$} & \textbf{PA-MPJPE$\downarrow$} & \textbf{MPJPE$\downarrow$} & \textbf{PA-MPJPE$\downarrow$} \\
\midrule
HMR~\cite{kanazawa2018end} & 393.30 & 124.15 & 103.36 (\textcolor{cvprblue}{\textbf{$\downarrow$74\%}}) & 58.79 (\textcolor{cvprblue}{\textbf{$\downarrow$53\%}}) \\
PARE~\cite{Kocabas2021PARE} & 364.49 & 120.06 & \textbf{83.73} (\textcolor{cvprblue}{\textbf{$\downarrow$77\%}}) & 63.16 (\textcolor{cvprblue}{\textbf{$\downarrow$47\%}}) \\
4D-Humans~\cite{goel2023humans} & 526.62 & 125.02 & 84.73 (\textcolor{cvprblue}{\textbf{$\downarrow$84\%}}) & \textbf{46.74} (\textcolor{cvprblue}{\textbf{$\downarrow$63\%}}) \\
\bottomrule
\end{tabular}
}
\caption{Comparison of 3D pose and shape estimation on InfantNet using adult body models (left) vs. infant body models (right). Metrics: MPJPE and PA-MPJPE (mm). Infant models consistently improve accuracy across all architectures, highlighting the importance of infant data and models.}
\label{tab:3dinfantnet_results}
\end{table}

\noindent\textbf{Results.}
We benchmark a wide range of 2D pose estimation models on InfantNet under two settings: (1) direct inference using COCO-pretrained weights, and (2) fine-tuning the models on our InfantNet training set. Table~\ref{tab:infantnet_results} summarizes the Average Precision (AP), Average Recall (AR), and their variants across 25 different configurations.

\textit{(a) Direct Inference Results.}
Sapiens achieves the highest AP (0.9008), while ViTPose-large and ViTPose-huge achieve the best AR (0.9199). HRNet-W48-Dark 256x192 is also competitive (AP: 0.8892, AR: 0.9069). 
Overall, transformer-based models, particularly Sapiens and ViTPose, show the best direct transfer to InfantNet.

\textit{(b) Fine-tuning on InfantNet.} Fine-tuning consistently boosts performance across models. On average, it yields a relative gain of \textbf{6.2\% in AP} and \textbf{5.2\% in AR} compared to direct inference. The ViTPose-huge model again achieves the best performance (AP: 0.9477, AR: 0.9582) after fine-tuning. Overall, models with higher input resolution outperform lower-resolution counterparts, demonstrating the importance of spatial detail for infant pose estimation.

\textit{(c) Source-specific and Cross-domain Evaluation.} We  evaluate model performance separately on the YouTube and HomeVisit subsets. For each source, we measure direct inference using COCO-pretrained weights, fine-tuning only on the InfantNet training portion of that source, and cross-domain generalization, where models are fine-tuned on one source (e.g., HomeVisit) and evaluated on the other (e.g., YouTube). Fine-tuning yields a relative gain of \textbf{6.1\% in AP} and \textbf{5.3\% in AR} on HomeVisit and \textbf{5.0\% in AP} and \textbf{3.7\% in AR} on YouTube. The source-specific and cross-domain comparison table is provided in Appendix Table~\ref{tab:youtube_results} and \ref{tab:HomeVisit_results}.

\begin{figure}[t]
    \centering
    \includegraphics[width=0.8\linewidth]{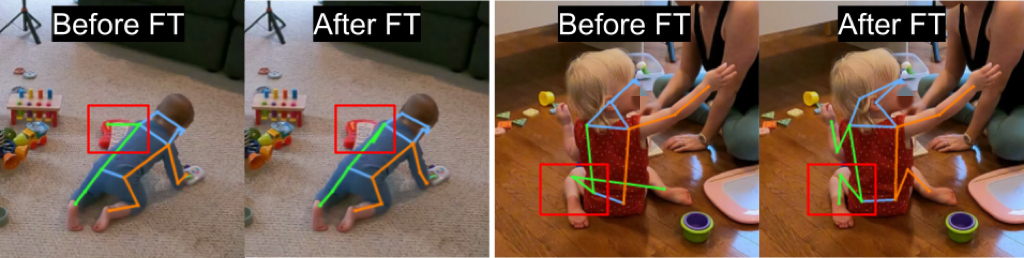}
    \caption{Failure modes and effects of fine-tuning: Without InfantNet FT, keypoint estimators frequently fail, e.g., by mislocalizing the left arm under severe self-occlusion (left) or predicting leg incorrectly in atypical infant poses (right). After FT, models can accurately localize joints in these challenging scenarios.}
    \label{fig:failure_mode}
    \vspace{-10pt}
\end{figure}

\noindent\textbf{Failure Mode Analysis.} Models without InfantNet fine-tuning fail on infant-specific poses (e.g., high flexibility), limb proportions (shorter limbs relative to torso), and self-occlusions (Fig.~\ref{fig:failure_mode}). After FT, models locate joints correctly.

\noindent\textbf{Hardest Joints.} Joint PCK statistics across body parts (Table~\ref{tab:vitpose_infantnet}) show that \textbf{leg} joints are the most challenging. Before fine-tuning, ViTPose-H achieves the lowest PCK on legs (0.793), compared to head (0.962), torso (0.842), and arms (0.865). Fine-tuning on InfantNet improves performance across all regions, including legs (0.793$\rightarrow$0.828), yet lower limbs remain challenging. This gap highlights challenges in infant pose estimation, including large articulation, different body proportions from adults, and frequent occlusions.

\begin{table}[t]
\centering
\setlength{\tabcolsep}{12pt}
\resizebox{\linewidth}{!}{%
\begin{tabular}{lccccc}
\toprule
\textbf{Method} & \textbf{All} & \textbf{Head} & \textbf{Torso} & \textbf{Arms} & \textbf{Legs} \\
\midrule
ViTPose-H (Before FT) & 0.871 & 0.962 & 0.842 & 0.865 & 0.793 \\
ViTPose-H (FT on InfantNet) & \textbf{0.906} & \textbf{0.986} & \textbf{0.890} & \textbf{0.901} & \textbf{0.828} \\
\bottomrule
\end{tabular}
}
\caption{PCK (\%) breakdown of ViTPose-H before and after fine-tuning on InfantNet. Fine-tuning improves all body parts, but leg joints remain the most challenging, reflecting the unique articulation and occlusion patterns in infant poses.}
\label{tab:vitpose_infantnet}
\vspace{-10pt}
\end{table}

\newcommand{\smilExpFig}{%
\begin{figure*}[t]
  \centering
  \includegraphics[width=1.0\linewidth]{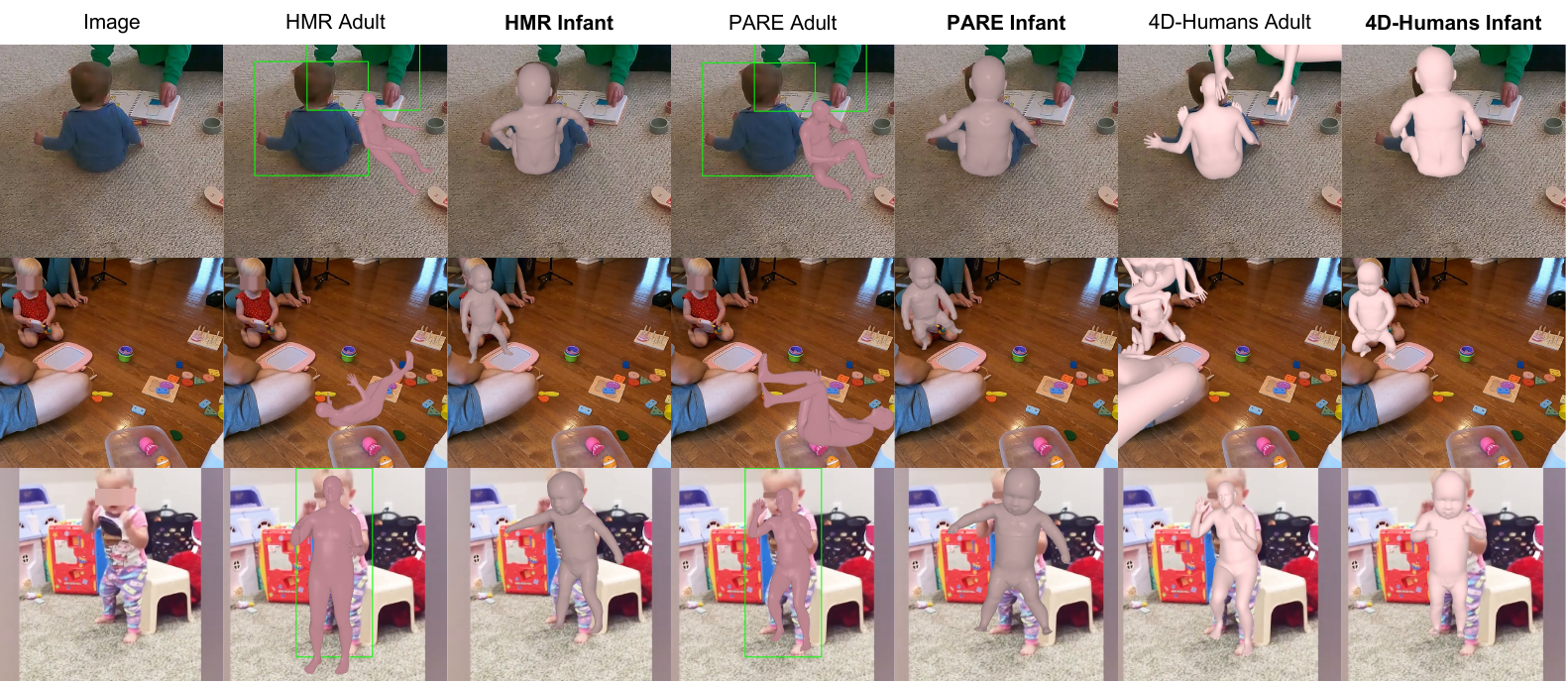}
  \caption{Visualization of 3D pose and shape estimation. From left to right, the columns show the input image, direct inference with adult body model and data, and training with infant body model and data for HMR~\cite{kanazawa2018end}, PARE~\cite{Kocabas2021PARE}, and 4D-Humans~\cite{goel2023humans}, respectively.}
  \label{fig:smil_exp_results}
\end{figure*}%
}
\ifcameraready\smilExpFig\fi

\subsection{3D Pose and Shape Estimation}
\label{subsec:3d_estimation}

We further benchmark representative 3D pose and shape estimation models on our proposed InfantNet dataset to assess their generalizability to infant body models, establishing baselines for 3D infant pose and shape estimation. 
Notably, a direct comparison with previous infant datasets is not possible, as InfantNet is the first to provide the high-quality 3D ground truth required for such evaluation (see Table~\ref{tab:related}).

\noindent\textbf{Setup.}
While no algorithms are explicitly designed for 3D infant pose and shape estimation, adult-based models can be adapted to infants. We benchmark CNN-based HMR~\cite{kanazawa2018end} and PARE~\cite{Kocabas2021PARE}, and transformer-based 4D-Humans~\cite{goel2023humans} as baselines. We provide a detailed analysis comparing end-to-end regression with SMIL-based optimization in Appendix~\ref{sec: end_opti}. We train each model for 100 epochs using a learning rate of \(2.5 \times 10^{-5}\) for HMR and PARE on 3 NVIDIA TITAN RTX GPUs, and a learning rate of \(2.0 \times 10^{-5}\) for 4D-Humans on 4 NVIDIA A5000 GPUs.

\noindent\textbf{Evaluation.}
We evaluate models with standard 3D human pose and shape estimation metrics. We report mean per joint position error (MPJPE) in mm, which measures the absolute error between predicted and ground-truth 3D joint positions, and Procrustes-aligned mean per joint position error (PA-MPJPE) in mm, which evaluates relative joint configuration error after rigid alignment.

\noindent\textbf{Model \& Data Preparation.}
We train the models on 10,071 frames and evaluate on 1,571 frames. HMR and 4D-Humans are trained with both 2D and 3D supervision. For PARE, we segment the mean SMIL mesh into seven parts with Blender~\cite{blender} for part-attention and mask supervision.

\ifcameraready\else\smilExpFig\fi

\noindent\textbf{Results.}
We benchmark 3D human pose and shape estimation models on InfantNet under two settings: (1) training and inference with SMIL-adapted weights, and (2) direct inference with pretrained adult-model weights. Fig.~\ref{fig:smil_exp_results} presents visualizations of both settings. Table~\ref{tab:3dinfantnet_results} reports MPJPE and PA-MPJPE across six configurations. 

\textit{(a) Adult Body Model.} Direct inference using human pose and shape estimation models yields lower accuracy compared to results obtained with infant-specific body models, affirming the domain gap between adult and infant poses and shapes. Among adult-centric models, PARE achieves the strongest performance (MPJPE: 364.49 mm, PA-MPJPE: 120.06 mm), while its accuracy remains considerably lower than when trained with infant body models.

\textit{(b) Infant Body Model.} Adapting adult-oriented models to SMIL substantially improves performance across all architectures. On average, there is a \textbf{78\%} reduction in MPJPE and \textbf{54\%} reduction in PA-MPJPE compared to direct adult-model inference. We provide cross-subset results in Appendix~\ref{sec: 3d_cross-subset}, demonstrating strong cross-domain generalization. Our infant-specific estimations are almost as good as adult-based estimations (e.g., HMR reports 87.97 mm MPJPE on Human3.6M). Among the models, 4D-Humans achieves the strongest results (MPJPE: 84.73 mm, PA-MPJPE: 46.74 mm), likely due to transformer-based global reasoning. In comparison, keypoint-only ZeDO-i~\cite{zhou2024zedo} achieves 97.1/77.3 mm. These results show our adapted models provide a better geometric prior for the infant body than adult-oriented models. To ensure privacy, digital facial masking is added. Additional experiments show negligible performance changes (approximately 1-5 mm) with masking, and details are provided in Appendix~\ref{sec: 3d_face_masking}. These adapted end-to-end models establish a new state-of-the-art for monocular 3D infant pose and shape estimation, providing a solid foundation for applications in developmental monitoring, childcare, and early medical diagnosis.

\subsection{Real-world Application}
\label{subsec:application}
To demonstrate InfantNet's practical impact, we evaluate its performance on detecting hand-to-mouth (HTM) behavior, a critical indicator of environmental exposure (\textit{e.g.}, soil/dust ingestion)~\cite{lupolt2025application, lupolt2025development}. Using 3D estimates from our framework, we identify HTM events via normalized hand-to-mouth Euclidean distances. Our infant-specific 4D-Humans model achieves \textbf{83\%} detection accuracy, compared to \textbf{30\%} of the adult-oriented baseline. This performance gap confirms that anatomical fidelity is not merely a refinement but a prerequisite for reliable infant behavior analysis and risk assessment in real-world settings.

\section{Conclusion}
\label{sec:conclusion}

We introduce InfantNet, the first large-scale dataset dedicated to infant pose and shape estimation, addressing the limitations of existing adult-centric models in this domain. Our results show that adult-trained models generalize poorly to infant morphology, while training on InfantNet effectively bridges this gap, enabling accurate 2D and 3D estimation. Beyond benchmarking, our framework provides a reliable foundation for downstream healthcare applications, such as quantifying infant hand-to-mouth behavior for exposure risk assessment.
While our current 3D subset provides a rigorous foundation, we aim to scale the dataset significantly by implementing a human-in-the-loop active learning framework. This will allow for the efficient expansion of high-quality 3D labels across more diverse domestic environments and ages.
Developed under strict ethical oversight with digital face-masking and controlled access, InfantNet is designed for responsible research. By releasing our data, models, and evaluation tools, we provide a standardized platform to advance infant-focused vision models, and ultimately foster positive impact in both academic research and pediatric care.

{
    \small
    \bibliographystyle{ieeenat_fullname}
    \bibliography{main}
}

\clearpage
\appendix
\setcounter{page}{1}
\maketitlesupplementary

\setcounter{section}{0}
\setcounter{table}{0}
\setcounter{figure}{0}

\renewcommand{\thetable}{S\arabic{table}}
\renewcommand{\thefigure}{S\arabic{figure}}

\section{Appendix}
\subsection{2D Source-Specific Evaluations}
\label{sec:2d_source_specific}
We analyze model performance by separately training and evaluating on the YouTube and HomeVisit subsets to understand how well existing human pose estimators adapt to each subset in isolation.

\paragraph{Setup. }
We follow the evaluation procedure outlined in \cref{subsec:2d_keypoint_estimation}, using the same model settings and metrics.

\paragraph{Evaluation. } 
For each source, models are fine-tuned on the corresponding training split 
(\(\text{YouTube}_{\text{Train}}\) or \(\text{HomeVisit}_{\text{Train}}\)) 
and evaluated on the matching test split 
(\(\text{YouTube}_{\text{Test}}\) or \(\text{HomeVisit}_{\text{Test}}\)). 
No cross-source data are used during training in this setting. 
All identity separations within each source are preserved. 
Metrics follow the standard COCO keypoint protocol over 17 keypoints: 
mean Average Precision (AP), AP.5, AP.75, and Average Recall (AR). 
We report results using the checkpoint that achieves the best validation AP on each source.

\paragraph{Result. }

\begin{table*}[t]
\centering
\resizebox{\linewidth}{!}{
\begin{tabular}{l|cccc|cccc|cccc}
\toprule
\textbf{Model} & \multicolumn{4}{c|}{\textbf{Direct Inference}} & \multicolumn{4}{c|}{\textbf{Source-Specific Fine-tuned}} & \multicolumn{4}{c}{\textbf{Cross-Subset Fine-tuned}}\\
\cmidrule(lr){2-5} \cmidrule(lr){6-9} \cmidrule(lr){10-13}
 & \textbf{AP} & \textbf{AP.5} & \textbf{AP.75} & \textbf{AR} & \textbf{AP} & \textbf{AP.5} & \textbf{AP.75} & \textbf{AR} &\textbf{AP} & \textbf{AP.5} & \textbf{AP.75} & \textbf{AR}\\
\midrule
Hourglass 256x256 & 0.8588	&0.9790	&0.9440	&0.8868	&0.9193	&0.9898	&0.9678 &0.9374 & 0.8667 &0.9896 &0.9566 &0.8962\\
Hourglass 384x384 & 0.8707	&0.9794	&0.9452	&0.8987	&0.9284	&0.9897	&0.9781	&0.9450 & 0.8808 &0.9894 &0.9566 &0.9091\\
HRFormer-small\_256x192 & 0.8761 &0.9779 &0.9537 &0.9062 &0.9247 &0.9896 &0.9778 &0.9422 & 0.8734 &0.9896 &0.9549 &0.9023\\
HRFormer-small\_384x288 & 0.8852 &0.9887 &0.9557 &0.9145 &0.9307 &0.9895 &0.9788 &0.9475 & 0.8816 &0.9896 &0.9658 &0.9115\\
HRFormer-base\_256x192 & 0.8900	&0.9883	&0.9559	&0.9172	&0.9325	&0.9893	&0.9782	&0.9489 & 0.8964 &0.9898 &0.9673 &0.9235\\
HRFormer-base\_384x288 & 0.8993	&0.9875	&0.9640	&0.9257	&0.9371	&0.9893	&0.9789	&0.9535 & 0.9063 &0.9897 &0.9681 &0.9301\\
Res50-256x192 & 0.8629	&0.9888	&0.9444	&0.8936	&0.9162	&0.9898	&0.9682	&0.9354 &0.8601	&0.9793	&0.9558	&0.8914\\
Res50-384x288 & 0.8583	&0.9781	&0.9424	&0.8906	&0.9228	&0.9900	&0.9790	&0.9399 &0.8658	&0.9894	&0.9458	&0.8958\\
Res101-256x192 & 0.8736	&0.9887	&0.9546	&0.9024	&0.9205	&0.9898	&0.9785	&0.9400 &0.8638	&0.9794	&0.9553	&0.8939\\
Res101-384x288 & 0.8720	&0.9782	&0.9543	&0.9007	&0.9301	&0.9898	&0.9788	&0.9453 &0.8832	&0.9798	&0.9556	&0.9096\\
Res152-256x192 & 0.8790	&0.9891	&0.9551	&0.9070	&0.9263	&0.9898	&0.9786	&0.9428 &0.8784	&0.9898	&0.9573	&0.9078\\
Res152-384x288 & 0.8864	&0.9889	&0.9556	&0.9132	&0.9319	&0.9897	&0.9784	&0.9480 &0.8913	&0.9896	&0.9564	&0.9175\\
HRNet-w32\_dark\_256x192 & 0.8888 &0.9788 &0.9567 &0.9151 &0.9312 &0.9894 &0.9784 &0.9470 & 0.8938	&0.9896	&0.9665	&0.9165\\
HRNet-w32\_dark\_384x288 & 0.8954 &0.9885 &0.9655 &0.9204 &0.9343 &0.9894 &0.9778 &0.9517 &0.8992	&0.9896	&0.9685	&0.9220\\
HRNet-w48\_dark\_256x192 & 0.8977 &0.9888 &0.9663 &0.9234 &0.9333 &0.9887 &0.9770 &0.9501 &0.8977	&0.9898	&0.9677	&0.9210\\
HRNet-w48\_dark\_384x288 & 0.8961 &0.9886 &0.9662 &0.9225 &0.9366 &0.9880 &0.9768 &0.9535 &0.9056	&0.9898	&0.9671	&0.9276\\
HRNet-w32\_256x192 & 0.8828	&0.9888	&0.9561	&0.9084	&0.9241	&0.9898	&0.9778	&0.9427 &0.8861	&0.9895	&0.9670	&0.9116\\
HRNet-w32\_384x288 & 0.8889	&0.9882	&0.9653	&0.9160	&0.9333	&0.9895	&0.9782	&0.9496 &0.8940	&0.9894	&0.9673	&0.9185\\
HRNet-w48\_256x192 & 0.8890	&0.9886	&0.9559	&0.9161	&0.9292	&0.9889	&0.9774	&0.9476 &0.8924	&0.9899	&0.9675	&0.9176\\
HRNet-w48\_384x288 & 0.8935	&0.9789	&0.9560	&0.9179	&0.9355	&0.9886	&0.9779	&0.9520 &0.8986	&0.9896	&0.9677	&0.9229\\
ViTPose-small\_256x192 & 0.8826	&0.9796	&0.9567	&0.9087	&0.9273	&0.9900	&0.9788	&0.9434 &0.8864	&0.9798	&0.9580	&0.9118\\
ViTPose-base\_256x192 & 0.9050	&0.9895	&0.9679	&0.9271	&0.9350	&0.9897	&0.9791	&0.9504 &0.9112	&0.9896	&0.9779	&0.9317\\
ViTPose-large\_256x192 & \textbf{0.9233}	&0.9886	&0.9760	&\textbf{0.9443}	&0.9443	&0.9895	&0.9785	&\textbf{0.9600} &0.9332 &0.9895 &0.9785 &0.9501\\
ViTPose-huge\_256x192 & 0.9184	&0.9876	&0.9660	&0.9437	&\textbf{0.9446}	&0.9887	&0.9780	&0.9597 &\textbf{0.9347}	&0.9888	&0.9779	&\textbf{0.9521}\\
\bottomrule
\end{tabular}
}
\caption{Comparison of 2D pose estimation models on the InfantNet Subsets (\(\text{YouTube}_{\text{Test}}\)) benchmark under three evaluation settings: direct inference on \(\text{YouTube}_{\text{Test}}\) using COCO-pretrained models (left block), source-specific fine-tuning on \(\text{YouTube}_{\text{Train}}\) (middle block), and cross-subset fine-tuning on \(\text{HomeVisit}_{\text{Train}}\) (right block). We report Average Precision (AP), AP at IoU thresholds 0.5 (AP.5) and 0.75 (AP.75), and Average Recall (AR). Across all architectures, fine-tuning substantially improves performance. 
On average, source-specific adaptation yields the relative gains of \textbf{5.0\% in AP} and \textbf{3.7\% in AR}, 
whereas cross-subset fine-tuning provides margin relative gains of
\textbf{0.5\% in AP} and \textbf{0.3\% in AR} over direct inference.
}
\label{tab:youtube_results}
\end{table*}

\begin{table*}[t]
\centering
\resizebox{\linewidth}{!}{%
\begin{tabular}{l|cccc|cccc|cccc}
\toprule
\textbf{Model} & \multicolumn{4}{c|}{\textbf{Direct Inference}} & \multicolumn{4}{c|}{\textbf{Source-Specific Fine-tuned}} & \multicolumn{4}{c}{\textbf{Cross-Subset Fine-tuned}}\\
\cmidrule(lr){2-5} \cmidrule(lr){6-9} \cmidrule(lr){10-13}
 & \textbf{AP} & \textbf{AP.5} & \textbf{AP.75} & \textbf{AR} & \textbf{AP} & \textbf{AP.5} & \textbf{AP.75} & \textbf{AR} & \textbf{AP} & \textbf{AP.5} & \textbf{AP.75} & \textbf{AR}\\
\midrule
Hourglass 256x256 & 0.8643	&0.9570	&0.9257	&0.8834	&0.9222	&0.9898	&0.9689 &0.9357 &0.9116	&0.9799	&0.9687	&0.9273\\
Hourglass 384x384 & 0.8724	&0.9572	&0.9358	&0.8907	&0.9288	&0.9897	&0.9790	&0.9415 &0.9177	&0.9797	&0.9689	&0.9329\\
HRFormer-small\_256x192 & 0.8766 &0.9573 &0.9359 &0.8954 &0.9235 &0.9896 &0.9687 &0.9369 &0.9194 &0.9796 &0.9685 &0.9350\\
HRFormer-small\_384x288 & 0.8774 &0.9568 &0.9352 &0.8960 &0.9261 &0.9893 &0.9688 &0.9398 &0.9219 &0.9896 &0.9688 &0.9369\\
HRFormer-base\_256x192 & 0.8769	&0.9560	&0.9346	&0.8977	&0.9326	&0.9897	&0.9791	&0.9452 &0.9260	&0.9893	&0.9688	&0.9423\\
HRFormer-base\_384x288 & 0.8832	&0.9559	&0.9343	&0.9023	&0.9369	&0.9895	&0.9788	&0.9487 &0.9294	&0.9893	&0.9688	&0.9447\\
Res50-256x192 & 0.8621	&0.9575	&0.9258	&0.8833	&0.9189	&0.9898	&0.9691	&0.9330 &0.9102	&0.9795	&0.9686	&0.9248\\
Res50-384x288 & 0.8654	&0.9573	&0.9256	&0.8848	&0.9212	&0.9899	&0.9691	&0.9346 &0.9150	&0.9797	&0.9686	&0.9294\\
Res101-256x192 & 0.8718	&0.9569	&0.9357	&0.8897	&0.9263	&0.9899	&0.9791	&0.9400 &0.9186	&0.9797	&0.9690	&0.9331\\
Res101-384x288 & 0.8763	&0.9569	&0.9356	&0.8941	&0.9291	&0.9899	&0.9693	&0.9425 &0.9246	&0.9896	&0.9691	&0.9398\\
Res152-256x192 & 0.8700	&0.9566	&0.9263	&0.8893	&0.9308	&0.9898	&0.9791	&0.9428 &0.9211	&0.9796	&0.9689	&0.9374\\
Res152-384x288 & 0.8778	&0.9569	&0.9354	&0.8955	&0.9338	&0.9899	&0.9791	&0.9449 &0.9246	&0.9799	&0.9691	&0.9389\\
HRNet-w32\_dark\_256x192 & 0.8844 &0.9575 &0.9363 &0.8999 &0.9329 &0.9898 &0.9791 &0.9449 &0.9256 &0.9797 &0.9689 &0.9398\\
HRNet-w32\_dark\_384x288 & 0.8831 &0.9568 &0.9355 &0.9017 &0.9365 &0.9899 &0.9794 &0.9475 &0.9267 &0.9895 &0.9690 &0.9426\\
HRNet-w48\_dark\_256x192 & 0.8849 &0.9571 &0.9362 &0.9013 &0.9369 &0.9898 &0.9794 &0.9485 &0.9278 &0.9797 &0.9691 &0.9415\\
HRNet-w48\_dark\_384x288 & 0.8815 &0.9570 &0.9356 &0.9005 &0.9383 &0.9897 &0.9794 &0.9500 &0.9317 &0.9896 &0.9789 &0.9450\\
HRNet-w32\_256x192 & 0.8788	&0.9576	&0.9361	&0.8965	&0.9294	&0.9899	&0.9792	&0.9426 &0.9191	&0.9797	&0.9686	&0.9340\\
HRNet-w32\_384x288 & 0.8797	&0.9571	&0.9355	&0.8979	&0.9327	&0.9898	&0.9790	&0.9436 &0.9222	&0.9796	&0.9687	&0.9380\\
HRNet-w48\_256x192 & 0.8809	&0.9572	&0.9357	&0.8978	&0.9336	&0.9897	&0.9790	&0.9454 &0.9211	&0.9797	&0.9690	&0.9374\\
HRNet-w48\_384x288 & 0.8801	&0.9568	&0.9356	&0.8992	&0.9369	&0.9897	&0.9792	&0.9478 &0.9274	&0.9894	&0.9691	&0.9428\\
ViTPose-small\_256x192 & 0.8714	&0.9568	&0.9357	&0.8898	&0.9244	&0.9898	&0.9690	&0.9378 &0.9190	&0.9895	&0.9690	&0.9344\\
ViTPose-base\_256x192 & 0.8849 &0.9566 &0.9361 &0.9029 &0.9348 &0.9898 &0.9790 &0.9473 &0.9252	&0.9892	&0.9688	&0.9407\\
ViTPose-large\_256x192 & \textbf{0.8901} &0.9560 &0.9350 &0.9117 &\textbf{0.9433} &0.9896 &0.9791 &\textbf{0.9567} &0.9347	&0.9893	&0.9784	&0.9498\\
ViTPose-huge\_256x192 & 0.8890 &0.9561 &0.9439 &\textbf{0.9120} &0.9413 &0.9897 &0.9791 &0.9561 &\textbf{0.9362}	&0.9893	&0.9783	&\textbf{0.9501}\\
\bottomrule
\end{tabular}
}
\caption{Comparison of 2D pose estimation models on the InfantNet Subsets(\(\text{HomeVisit}_{\text{Test}}\)) benchmark under three evaluation settings: direct inference using COCO-pretrained models (left block), fine-tuning on \(\text{HomeVisit}_{\text{Train}}\) (middle block), and fine-tuning on \(\text{YouTube}_{\text{Train}}\)(right block). We report Average Precision (AP), AP at IoU thresholds 0.5 (AP.5) and 0.75 (AP.75), and Average Recall (AR). Across all architectures, fine-tuning substantially improves performance. 
In average, source-specific adaptation yields the  relative gains of \textbf{6.1\% in AP} and \textbf{5.3\% in AR}, 
whereas cross-subset fine-tuning offers comparably strong relative gains of  
\textbf{5.2\% in AP} and \textbf{4.7\% in AR} over direct inference.
}
\label{tab:HomeVisit_results}
\end{table*}

In addition to the mixed-subset results presented in the main paper, we further
benchmark the models on each source independently. Specifically, we evaluate
their performance on the YouTube and HomeVisit subsets under two settings:
(1) direct inference using COCO-pretrained weights, and (2) fine-tuning on the
source-specific training split. Table~\ref{tab:youtube_results} and~\ref{tab:HomeVisit_results}'s left two blocks report the corresponding
Average Precision (AP), Average Recall (AR), AP.5, and AP.75 for all evaluated
configurations.

\textit{(a) Direct Inference on YouTube Test Set.}
On the \textbf{YouTube Test Set}, ViTPose-large achieves the highest performance among all
models, obtaining an AP of 0.9233 and an AR of 0.9443.
ViTPose-huge also performs strongly (AP: 0.9184, AR: 0.9437). Among CNN-based models,
HRNet-W48-Dark\_256x192 is the strongest performer (AP: 0.8977, AR: 0.9234),
while the transformer-based HRFormer-base\_384x288 achieves AP and AR of 0.8993 and 0.9257,
respectively. 

\textit{(b) Direct Inference on HomeVisit Test Set.}
On the \textbf{HomeVisit Test Set}, ViTPose-large again achieves the highest AP
(0.8901) and ViTPose-huge achieves the highest AR (0.9120). CNN models perform competitively,
with HRNet-W48-Dark\_256x192 reaching AP(0.8849) and
AR(0.9013). Across both subsets, transformer-based ViTPose models consistently
achieve the strongest direct-inference results, and larger model variants
within each family outperform their smaller counterparts.

\textit{(c) Fine-tuning on YouTube Training Set.}
Fine-tuning on the YouTube training subset consistently improves performance
across all architectures. On average, fine-tuning yields a relative gain of
\textbf{5.0\% in AP} and \textbf{3.7\% in AR} compared to direct inference.
Among transformer-based models, ViTPose-huge achieves the highest AP (0.9446) after
adaptation, while ViTPose-large obtains the highest AR
(0.9600). CNN-based models also benefit substantially from
fine-tuning; HRNet-W48-Dark\_384x288 improves from an AP of 0.8961
to 0.9366 and from an AR of 0.9225 to 0.9535.

\textit{(d) Fine-tuning on HomeVisit Training Set.} Fine-tuning on the HomeVisit training subset yields consistent improvements across all architectures. On average, fine-tuning provides a relative gain of
\textbf{6.1\% in AP} and \textbf{5.3\% in AR} compared to direct inference.
ViTPose-large achieves the highest AP and AR after adaptation (AP: 0.9433, AR: 0.9567). CNN-based models also show substantial gains; Res152\_256x192 improves from an AP of 0.8700 to 0.9308 and from an AR of 0.8893 to 0.9428.

\subsection{2D Cross-Subset Evaluations}
\label{sec: 2d_cross-subset}
We next evaluate cross-subset generalization by fine-tuning on one source and testing on the other.  This setting allows us to assess the robustness of each model under domain shift. It also reveals how well knowledge learned from one environment transfers to a visually distinct target domain.

\paragraph{Setup. } The experiment uses the same protocol detailed in \cref{subsec:2d_keypoint_estimation}, maintaining identical model configurations and evaluation metrics. 

\paragraph{Evaluation. } For cross-subset evaluation, models are fine-tuned on the training split of one
source and evaluated on the test split of the other. Specifically, we consider
two cross-subset directions: models fine-tuned on
\(\text{HomeVisit}_{\text{Train}}\) are evaluated on \(\text{YouTube}_{\text{Test}}\),
and models fine-tuned on \(\text{YouTube}_{\text{Train}}\) are evaluated on
\(\text{HomeVisit}_{\text{Test}}\). No data from the target source are used
during training. Identity separations within each source are preserved.
Evaluation in this section adopts the same COCO keypoint metrics and checkpoint
selection protocol introduced in our source-specific experiments, ensuring methodological consistency
across all experiments.

\paragraph{Result. } In addition to the source-specific results reported in \cref{sec:2d_source_specific}, we further evaluate cross-subset generalization between the YouTube and HomeVisit subsets.
Table~\ref{tab:youtube_results} and~\ref{tab:HomeVisit_results}'s rightmost blocks report the corresponding Average Precision (AP), Average Recall (AR), AP.5, and AP.75 for all configurations.

\textit{Fine-tuning on HomeVisit and Evaluating on YouTube.}
Fine-tuning on the HomeVisit subset produces only marginal gains on the YouTube test set, with an average relative gain of \textbf{0.5\% in AP} and \textbf{0.3\% in AR}. This limited increase is expected and reflects a clear domain mismatch: HomeVisit videos feature fixed viewpoints, stable lighting, and clean indoor scenes, whereas YouTube videos contain far greater variability in camera motion, compression artifacts, occlusions, and infant appearance. Consequently, models fine-tuned on HomeVisit adapt to a simpler and narrower distribution, offering little transfer to the more complex YouTube subset. Higher-capacity models still show modest improvements. ViTPose-huge reaches the highest AP (0.9347)  and the highest AR (0.9521) after fine-tuning.

\textit{Fine-tuning on YouTube and Evaluating on HomeVisit.}
Fine-tuning on the YouTube train set leads to substantial improvements on the HomeVisit test set, with an average relative gain of \textbf{5.2\% in AP} and \textbf{4.7\% in AR}. It reflects the fact that YouTube provides a larger and more diverse training distribution that transfers effectively to the more controlled HomeVisit subset. All model families benefit consistently from this direction of adaptation. Transformer-based architectures exhibit the largest absolute performance, with ViTPose-large and ViTPose-huge reaching AP values of 0.9347 and 0.9362, respectively, after fine-tuning. Overall, these results demonstrate a clear asymmetry in cross-subset transfer: training on the more complex and diverse YouTube domain significantly boosts performance on the HomeVisit subset, whereas the reverse direction yields only marginal gains. This highlights the value of high-variability data for robust infant pose estimation.

\begin{figure*}
  \centering
  \includegraphics[width=0.82\linewidth]{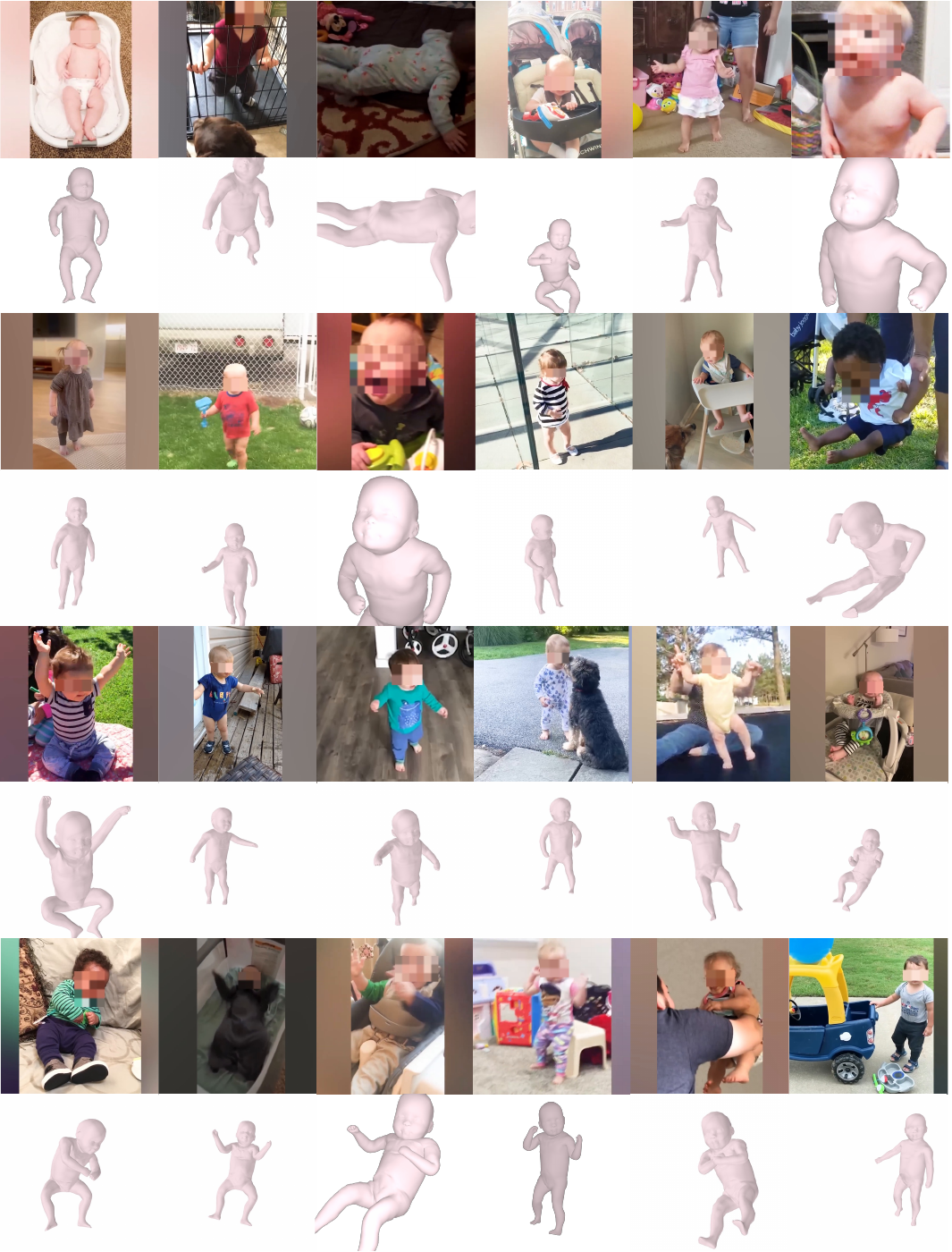}
  \caption{Additional examples of meshes generated via optimization-based SMIL fitting for the InfantNet dataset. The refined SMIL parameters become the 3D annotations used to train end-to-end infant models. InfantNet captures a diverse set of infant shapes and poses with high-quality 2D and 3D annotations.}
  \label{fig:3d_annotations_additional_samples}
\end{figure*}

\subsection{3D Infant Pose and Shape Annotations}
\label{sec: 3d_annotations}
Figure~\ref{fig:3d_annotations_additional_samples} presents additional examples of the 3D infant meshes obtained through parameters generated from our SMIL optimization process. These visualizations illustrate the diversity of infant poses and body shapes captured in the fitted meshes used as 3D annotations in InfantNet to train the end-to-end models, complementing the samples shown in the main paper.

\begin{table}[t]
\centering
\setlength{\tabcolsep}{12pt} 
\begin{tabular}{l|cc}
\toprule
\textbf{Model} & \textbf{MPJPE$\downarrow$} & \textbf{PA-MPJPE$\downarrow$} \\
\midrule
HMR~\cite{kanazawa2018end} & 118.53 & 62.30 \\
PARE~\cite{Kocabas2021PARE} & 107.52 & 63.83 \\
4D-Humans~\cite{goel2023humans} & \textbf{84.50} & \textbf{42.40} \\
\bottomrule
\end{tabular}
\caption{Cross-subset 3D pose and shape estimation. The models are trained on the HomeVisit subset and tested on the YouTube subset. Metrics: MPJPE and PA-MPJPE (mm). Infant models consistently improve accuracy across all architectures, even with cross-subset training and evaluation, validating the generalizability of the infant models.}
\label{tab:3dcross_results}
\end{table}

\subsection{Occlusion and Edge Case Analysis}
\label{sec: edge_cases}
\begin{figure*}
  \centering
  \includegraphics[width=0.82\linewidth]{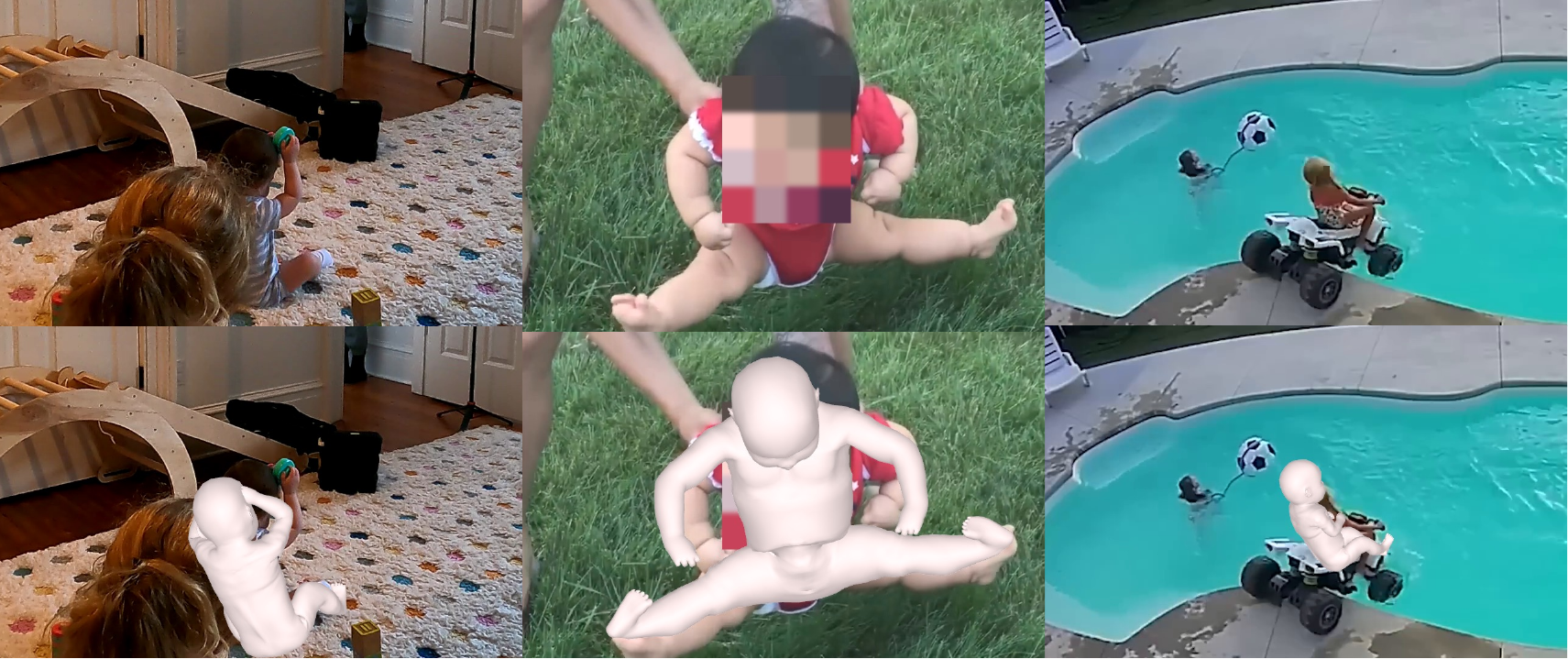}
  \caption{Examples of occlusion, pose, viewpoint edge cases, respectively.}
  \label{fig:edge_cases}
\end{figure*}

On frames with $\geq5$ occluded keypoints, the infant-specific 4D-Humans achieves \textbf{91.28} mm MPJPE and \textbf{48.84} mm PA-MPJPE, comparable to Table~\ref{tab:3dinfantnet_results}. We further define rare poses and viewpoints by 5th-NN distance in pelvis-centered SMIL-joint space and normalized camera-pose space, respectively. On the top 10\% rare-pose samples, the model achieves \textbf{103.53}/\textbf{57.25} mm MPJPE/PA-MPJPE, and \textbf{85.47}/\textbf{46.28} mm on the top 10\% rare-viewpoint samples, indicating robustness to challenging occlusions, poses, and viewpoints. Figure~\ref{fig:edge_cases} presents examples of our infant-specific 4D-Humans model on edge cases, including occlusion, rare poses, and rare viewpoints. 

\subsection{3D Cross-Subset Evaluations}
\label{sec: 3d_cross-subset}
We evaluate the generalizability of the 3D infant pose and shape estimation models under cross-subset conditions, where the HomeVisit subset serves as training data and the YouTube subset is used for evaluation. This experiment mimics a more realistic scenario and reflects the sensitivity of the infant models to variations in environmental settings. 
\paragraph{Setup. }
We follow the same training and evaluation protocol described in \cref{subsec:3d_estimation}, with identical model configurations and evaluation metrics.
\paragraph{Data Preparation. }
We train the models exclusively on the HomeVisit subset and evaluate on the YouTube subset to assess cross-subset generalization. The models are trained with \textbf{9,362 frames} and evaluated with \textbf{2,280 frames}.
\paragraph{Results. }
As shown in Table~\ref{tab:3dcross_results}, training and evaluating on different data sources leads to moderate accuracy drops compared to the results in \cref{subsec:3d_estimation}. Nevertheless, all infant models still outperform their adult baselines across architectures, indicating strong cross-domain generalization. Among them, 4D-Humans achieves the best performance (MPJPE: 84.50 mm, PA-MPJPE: 42.40 mm). The results suggest that the 3D infant pose and shape annotations of InfantNet retain cross-subset robustness despite differences in capture environmental settings.

\subsection{End-to-End vs Optimization}
\label{sec: end_opti}
To evaluate the benefits of learning-based 3D infant pose and shape estimation, we compare our end-to-end models from \cref{subsec:3d_estimation} against the optimization-based SMIL fitting process used to generate ground-truth 3D annotations in \cref{subsec:3d_fitting}. This comparison highlights two key aspects for end-to-end models: inference efficiency and robustness to cases where optimization fails or produces severely deformed meshes. By evaluating inference speed and estimation quality, we demonstrate that end-to-end regression not only achieves faster inference, but also yields more reliable estimations, reflecting the practical values of end-to-end models for 3D infant modeling.
\paragraph{Inference Speed. }
From Table~\ref{tab:3d_inference_speed}, end-to-end models are significantly faster than optimization-based SMIL fitting in terms of inference. While optimization requires an average of 38.7020 seconds per image for inference, end-to-end models operate in the millisecond range, with HMR achieving the fastest speed ($\approx3.25$ ms). The results clearly demonstrate the efficiency of end-to-end regression, enabling real-time inference that optimization-based methods cannot achieve.

\begin{figure}[t]
  \centering
  \includegraphics[width=0.54\linewidth]{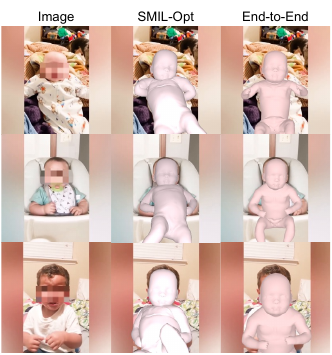}
  \caption{Visualization of 3D pose and shape estimation model performances. The columns from left to right: the input image, optimization-based SMIL fitting, and inference with end-to-end model (4D-Humans~\cite{goel2023humans}) trained with infant body model and data, respectively. The end-to-end 4D-Humans model produces anatomically plausible meshes that better align with the infant silhouette, while optimization-based SMIL fitting produces deformed meshes, illustrating the robustness of end-to-end models in 3D infant estimation compared to optimization-based fitting.}
  \label{fig:3d_inference_quality}
\end{figure}

\paragraph{Estimation Quality. }
We conduct this experiment to evaluate the reliability of end-to-end regression in cases where optimization-based SMIL fitting fails. As shown in Figure~\ref{fig:3d_inference_quality}, optimization outputs exhibit mesh distortions, such as implausible limb articulations and inconsistent body proportions. In contrast, the end-to-end 4D-Humans model produces anatomically coherent meshes that align better with the underlying infant silhouette. The results highlight the practical significance of learning-based regression, yielding more reliable 3D infant pose and shape estimations.

\begin{table}
\centering
\setlength{\tabcolsep}{18pt} 
\begin{tabular}{l c}
\toprule
\textbf{Model} & \textbf{Avg Time (s / img)} \\
\midrule
\textbf{Optimization} & \\
\quad SMIL-Opt~\cite{bogo2016keep,yin2025easyret3d} & 38.7020 \\
\midrule
\textbf{End-to-End Models} & \\
\quad HMR~\cite{kanazawa2018end} & \textbf{0.0033} \\
\quad PARE~\cite{Kocabas2021PARE} & 0.0091 \\
\quad 4D-Humans~\cite{goel2023humans} & 0.0212 \\
\bottomrule
\end{tabular}
\caption{Comparison of average inference speed between optimization-based SMIL fitting and end-to-end infant modeling models. End-to-end models achieve millisecond-level inference, demonstrating the substantial computational advantage of learning-based 3D infant pose and shape estimation models.}
\label{tab:3d_inference_speed}
\end{table}

\begin{table}[t]
\centering
\setlength{\tabcolsep}{12pt} 
\begin{tabular}{l|cc}
\toprule
\textbf{Model} & \textbf{MPJPE$\downarrow$} & \textbf{PA-MPJPE$\downarrow$} \\
\midrule
HMR~\cite{kanazawa2018end} & 100.00 & 55.60 \\
PARE~\cite{Kocabas2021PARE} & \textbf{78.48} & 59.41 \\
4D-Humans~\cite{goel2023humans} & 86.57 & \textbf{50.15} \\
\bottomrule
\end{tabular}
\caption{3D pose and shape estimation on masked infant inputs. Metrics: MPJPE and PA-MPJPE (mm). Applying masking results in negligible degradation, demonstrating model robustness and dataset viability.}
\label{tab:3dmask_results}
\end{table}

\subsection{3D Evaluation with Face-Masked Data}
\label{sec: 3d_face_masking}
We evaluate the impact of privacy-preserving face masking on 3D infant pose and shape estimation. This experiment assesses the robustness of the proposed end-to-end infant models to masked inputs and the research viability of our anonymized dataset for accurate 3D infant pose and shape estimation.
\paragraph{Setup. }
We again follow the same train and evaluation protocol described in \cref{subsec:3d_estimation}, using identical model configurations and evaluation metrics.
\paragraph{Data Preparation. }
We train and evaluate the models with the same set of data described in \cref{subsec:3d_estimation}, replacing the input images with versions in which the face regions are masked using our privacy-preserving pipeline. The masking removes identifiable facial features such as eyes and noses while preserving the overall appearance of the head and body contours, thereby retaining the visual cues for accurate 3D estimation. 
\paragraph{Results. }
As shown in Table~\ref{tab:3dmask_results}, masking-based anonymization does not introduce systematic degradation in 3D infant pose and shape estimation, with performance variations limited to approximately 1–5 mm compared to the original results in \cref{subsec:3d_estimation}. HMR and PARE exhibit slight improvements under masked inputs (e.g., PARE improves from 83.73 mm to 78.48 mm in MPJPE), while 4D-Humans shows a marginal decrease (84.73 mm to 86.57 mm in MPJPE). The improvements observed for the CNN-based models (HMR, PARE) may stem from their reliance on structural cues, and suppressing high-frequency facial details removes visual noise without affecting overall geometry. In contrast, the small performance drop for 4D-Humans suggests the sensitivity to fine-grained local texture cues of attention-based architecture. The minimal impact of anonymization indicates that the masking procedure preserves essential structural and geometric cues required for accurate 3D reconstruction, while effectively removing identifiable facial information. These findings validate both the robustness of our models and the viability of our privacy-preserving dataset design.

\subsection{Public Release}
\label{sec: public_release}
\paragraph{Data. }
To support future research in 3D infant modeling, we plan to release all the annotations of the training and evaluation data, with strict privacy safeguards upon acceptance of this paper, and checkpoints of the models fine-tuned on our InfantNet.
\paragraph{Code. }
We will release the implementation of our data processing and end-to-end 3D infant pose and shape estimation models for reproducibility.

\end{document}